\documentclass[10pt,twocolumn,letterpaper]{article}

\usepackage[pagenumbers]{cvpr} 

\usepackage[accsupp]{axessibility}  

\usepackage{algorithm}
\usepackage{algorithmic}

\usepackage{bm}
\usepackage{amsfonts}
\usepackage{amsmath}
\usepackage{booktabs}
\usepackage{multicol}
\usepackage{multirow}
\usepackage{amssymb}
\usepackage{makecell}

\definecolor{cvprblue}{rgb}{0.21,0.49,0.74}
\usepackage[pagebackref,breaklinks,colorlinks,allcolors=cvprblue]{hyperref}

\def\paperID{12422} 
\def\confName{CVPR}
\def\confYear{2025}

\title{Towards Training-free Anomaly Detection with Vision and Language Foundation Models}

\author{Jinjin Zhang\textsuperscript{1,2} \hspace{0.5em} 
Guodong Wang\textsuperscript{1,2} \hspace{0.5em}
Yizhou Jin\textsuperscript{2} \hspace{0.5em} 
Di Huang\textsuperscript{1,2}\footnotemark[1]  \\
\textsuperscript{1}State Key Laboratory of Complex and Critical Software Environment, \\Beihang University, Beijing 100191, China \\
\textsuperscript{2}School of Computer Science and Engineering, Beihang University, Beijing 100191, China\\
{\tt\small \{jinjin.zhang, wanggd, yizhou.jin, dhuang\}@buaa.edu.cn}
}

\begin{document}
\maketitle

\footnotetext[1]{Corresponding author.}

\begin{abstract}
Anomaly detection is valuable for real-world applications, such as industrial quality inspection. 
However, most approaches focus on detecting local structural anomalies while neglecting compositional anomalies incorporating logical constraints. 
In this paper, we introduce \textbf{LogSAD}, a novel multi-modal framework that requires no training for both \textbf{Log}ical and \textbf{S}tructural \textbf{A}nomaly \textbf{D}etection. 
First, we propose a match-of-thought architecture that employs advanced large multi-modal models (i.e. GPT-4V) to generate matching proposals, formulating interests and compositional rules of thought for anomaly detection. 
Second, we elaborate on multi-granularity anomaly detection, consisting of patch tokens, sets of interests, and composition matching with vision and language foundation models. 
Subsequently, we present a calibration module to align anomaly scores from different detectors, followed by integration strategies for the final decision.
Consequently, our approach addresses both logical and structural anomaly detection within a unified framework and achieves state-of-the-art results without the need for training, even when compared to supervised approaches, highlighting its robustness and effectiveness.
Code is available at \href{https://github.com/zhang0jhon/LogSAD}{https://github.com/zhang0jhon/LogSAD}.
\end{abstract}

\section{Introduction}

\begin{figure}[t]
\centering
\includegraphics[width=0.96\linewidth]{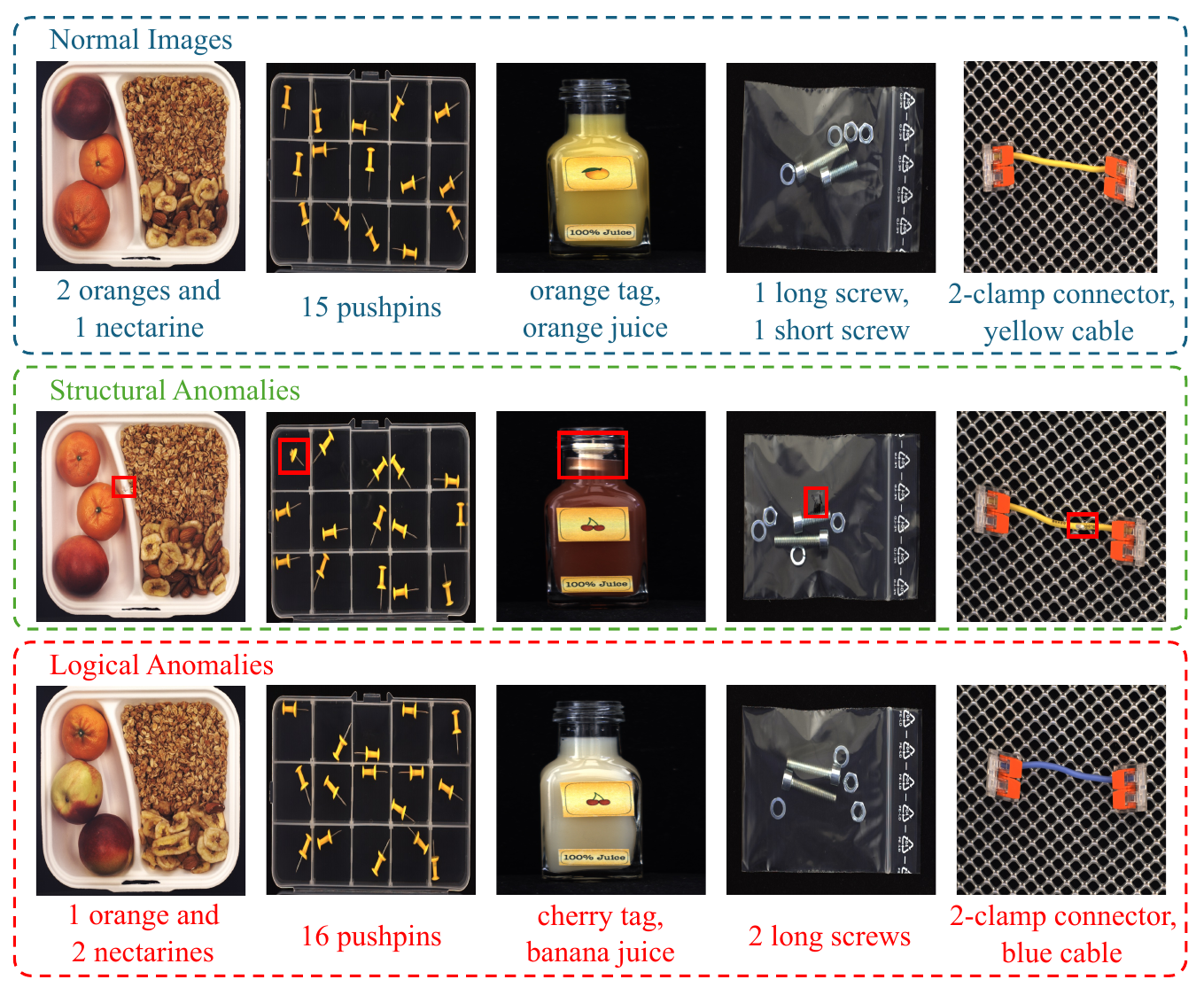} 
\caption{Examples of structural and logical anomalies in MVTec LOCO dataset~\cite{bergmann2022beyond}. Compositional multi-modal feature matching plays a crucial role in unified anomaly detection, particularly in identifying and categorizing logical anomalies effectively. }
\label{fig:mvtec_loco}
\end{figure}

Anomaly detection is widely employed in real-world applications, particularly in industrial quality inspection, to identify anomalies with limited normal data~\cite{golan2018deep, roth2022towards, li2024promptad, batzner2024efficientad}.
Existing anomaly detection methods have demonstrated impressive performance on anomaly detection datasets such as MVTec AD~\cite{bergmann2019mvtec} and VisA~\cite{zou2022spot}, which are biased towards local structural anomalies like scratches, dents, or contaminations.
However, those anomaly detection methods often fail to detect logical anomalies, such as incorrect wiring of circuits, permissible objects occurring in invalid locations, or the absence of essential components~\cite{bergmann2022beyond}.
To address this issue, several approaches have been proposed for logical anomaly detection~\cite{guo2023template, kim2024few}, yielding decent performance through accurate mask annotations and training efforts.

Despite considerable efforts in logical anomaly detection, many approaches focus primarily on the visual modality, which is necessary but insufficient for detecting high-level anomalies with logical constraints and often lack the ability to identify compositional aspects.
Furthermore, the need for precise annotations and intricate architectures complicates the practical applications in realistic scenarios, especially when addressing both structural and logical anomalies.
As illustrated in \cref{fig:mvtec_loco}, although visual feature matching is predominant in structural anomaly detection, compositional multi-modal feature matching is essential for distinguishing high-level logical anomalies, including attributes, entity-relationships, and other complex compositions.
Recent advancements in vision and language models (VLMs), notably CLIP~\cite{radford2021learning}, have highlighted that fine-tuning VLMs significantly impacts zero-shot and few-shot anomaly detection~\cite{jeong2023winclip, gu2024anomalygpt, zhou2024anomalyclip}.
However, while these VLM-based approaches are effective in detecting structural anomalies, they continue to face challenges with compositionality in logical anomalies~\cite{ma2023crepe, li2024promptad}.
Moreover, the potential of VLMs to concurrently detect both structural and logical anomalies remains largely unexplored.

In this paper, we introduce LogSAD, a unified multi-modal framework designed for both structural and logical anomaly detection without training endeavors. 
Firstly, we propose match-of-thought, utilizing advanced and powerful GPT-4V~\cite{achiam2023gpt} to generate matching proposals, formulating interests and compositional matching rules of thought with vision and language instructions. 
Secondly, we employ multiple anomaly detectors to detect anomalies across various granularities using vision and language foundation models, such as CLIP~\cite{radford2021learning} and SAM~\cite{kirillov2023segment}. 
Finally, we calibrate and fuse anomaly scores from different detectors to make final decisions within the unified framework.
Extensive experiments are conducted across various anomaly detection datasets to validate the effectiveness and robustness of our method.

The main contributions are summarized as follows:
\begin{itemize}
    \item We present LogSAD, a training-free framework for anomaly detection utilizing vision and language foundation models, and demonstrate its capability to detect both logical and structural anomalies.
    \item We introduce the match-of-thought architecture, illustrating its effectiveness through intermediate steps in generating interests and matching rules for anomaly detection with multi-modal instructions.
    \item We propose multi-granularity detectors encompassing patch tokens, set of interests, and composition matching, as well as fusion strategies within the unified framework for anomaly detection.
\end{itemize}

\section{Related Work}

\noindent\textbf{Vision and Language Foundation Models.} 
The past several years have witnessed a significant advancement in vision and language foundation models, as evidenced by~\cite{radford2021learning, oquab2023dinov2, kirillov2023segment, achiam2023gpt, team2023gemini, liu2024visual, li2024mini}. 
These foundation models demonstrate significant capability across various realistic scenarios, including zero-shot classification, open-vocabulary perception, and multi-modal learning.
SAM~\cite{kirillov2023segment} is among the leading vision foundation models, making substantial strides in zero-shot image segmentation. 
CLIP~\cite{radford2021learning} is the pioneering model to undertake vision and language pre-training on large-scale image-text pairs, demonstrating unprecedented generality in downstream tasks.
LLaVA~\cite{liu2024visual, liu2024improved} introduces an end-to-end trained large multi-modal model (LMM) that integrates a vision encoder with large language model (LLM) for comprehensive visual and language understanding.
GPT-4V~\cite{achiam2023gpt}, a large-scale multi-modal model, can accept both image and text inputs and generate text outputs, exhibiting human-level performance across a variety of professional and academic benchmarks.
Consequently, the utilization of vision and language foundation models has become ubiquitous in real-world applications, providing robustness and generalization in semantic and spatial understanding~\cite{wang2024sam}, multi-modal alignment~\cite{yu2022coca, li2023blip}, \emph{etc}.
Nevertheless, recent studies indicate that even the most advanced LMMs still face challenges in capturing aspects of compositionality in visual reasoning, such as attributes and relationships between objects~\cite{ma2023crepe, mitra2024compositional, zeng2024investigating}.

\begin{figure*}[t]
\centering
\includegraphics[width=0.96\textwidth]{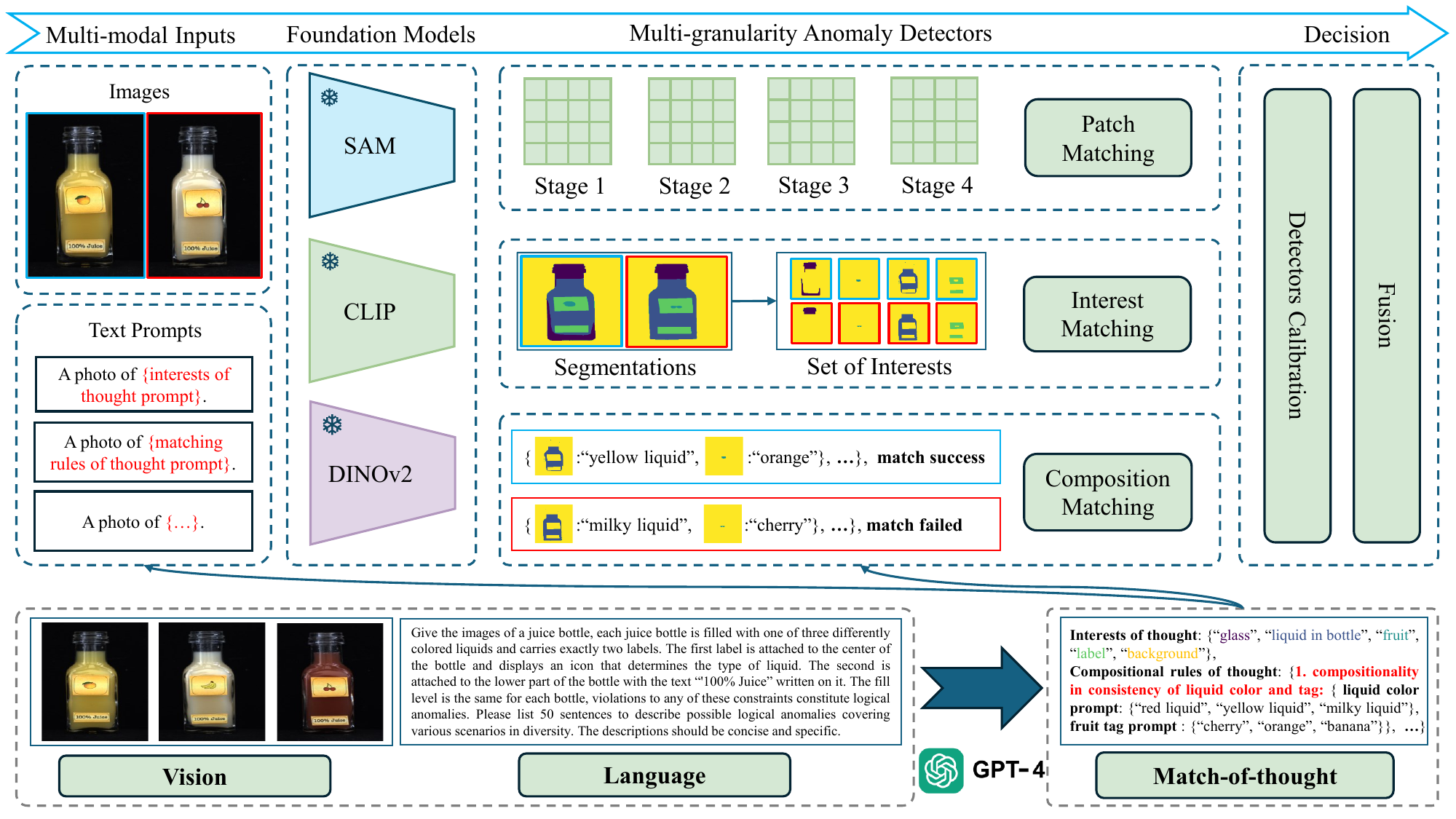} 
\caption{The framework of LogSAD. In the framework, we utilize match-of-thought to generate matching proposals, deriving text prompts of interests and compositional rules for anomaly detection. Based on the text prompts, our method leverages vision and language foundation models to achieve multi-granularity anomaly detection, followed by calibration and fusion modules to make final decision. Importantly, our algorithm detects both structural and logical anomalies within a unified framework, eliminating the need for training efforts. }
\label{fig:framework}
\end{figure*}

\noindent\textbf{Anomaly Detection.} 
Due to the scarcity of anomalies, most methods focus on anomaly detection with several normal images.
GCAD~\cite{bergmann2022beyond} introduces a new method for the unsupervised localization of anomalies which consists of two main branches, one of which is primarily responsible for the localization of structural anomalies and the other one for the localization of logical anomalies.
PSAD~\cite{kim2024few} focuses on logical anomaly detection and introduces a novel component segmentation model that leverages segment annotations of labeled images and unlabeled images sharing logical constraints. 
LogiCode~\cite{zhang2024logicode} introduces additional annotations in LOCO-Annotations dataset and LogiBench benchmark, addressing automatic code generation with LLMs for logical AD. 
PatchCore~\cite{roth2022towards} leverages a maximally representative memory bank of nominal patch features for structural anomaly detection.
UniAD~\cite{you2022unified} and OmniAL~\cite{zhao2023omnial} conduct structural anomaly detection across multiple categories using a unified framework.
WinCLIP~\cite{jeong2023winclip} proposes a window-based CLIP approach with compositional ensemble on state words and prompt templates, aiming for efficient extraction and aggregation of multi-level features.
AnomalyGPT~\cite{gu2024anomalygpt} explores the utilization of large vision-language models to address the industrial structural anomaly detection problem.
THFR~\cite{guo2023template} proposes a template-guided hierarchical feature restoration framework for anomaly detection, incorporating bottleneck compression and template-guided compensation for anomaly-free feature restoration.
PromptAD~\cite{li2024promptad} presents a one-class prompt learning method for few-shot anomaly detection, achieving the state-of-the-art performances on structural anomaly detection but still struggling with the challenging logical anomalies.
EfficientAD~\cite{batzner2024efficientad} exploits a teacher-student model and addresses the detection of logical anomalies with auto-encoder.
SimpleNet~\cite{liu2023simplenet} generates synthetic anomalies in a pretrained feature space to train a discriminator network for detecting anomalous features.
GRAD~\cite{dai2024generating} proposes a diffusion model, Patchdiff, to generate diverse contrastive images, and trains lightweight detectors for anomaly detection.
GeneralAD~\cite{strater2024generalad} proposes a self-supervised anomaly generation module to construct pseudo-abnormal samples, and employs a transformer-based discriminator capable of detecting a wide range of anomalies.

\section{Methodology}

In this section, we elaborate on the details of LogSAD, illustrating how it works for both logical and structural anomaly detection within a unified framework. As shown in \cref{fig:framework}, our method is built upon match-of-thought and employs multi-granularity detectors for anomaly detection with vision and language foundation models. The details are presented as follows. 

\subsection{Match-of-thought}
\label{sec:match-of-thought}

In contrast to structural anomalies that occur as scratches or dents in manufactured products, logical anomalies violate the underlying constraints of compositionality in vision and natural language, composed of visual and textural atoms (\emph{e.g.} objects in images or words in a sentence). 
Considering that LMMs struggle with compositional visual reasoning, we propose the match-of-thought (MoT), which involves a series of intermediate reasoning steps for prompt and match engineering in anomaly detection. 
Inspired by chain-of-thought approaches~\cite{wei2022chain, zhang2023multimodal}, the MoT extracts interests of thought from vision and language instructions, simultaneously formulating matching rules for different types of compositional logical anomalies.

\begin{figure}[ht]
\centering
\includegraphics[width=0.95\linewidth]{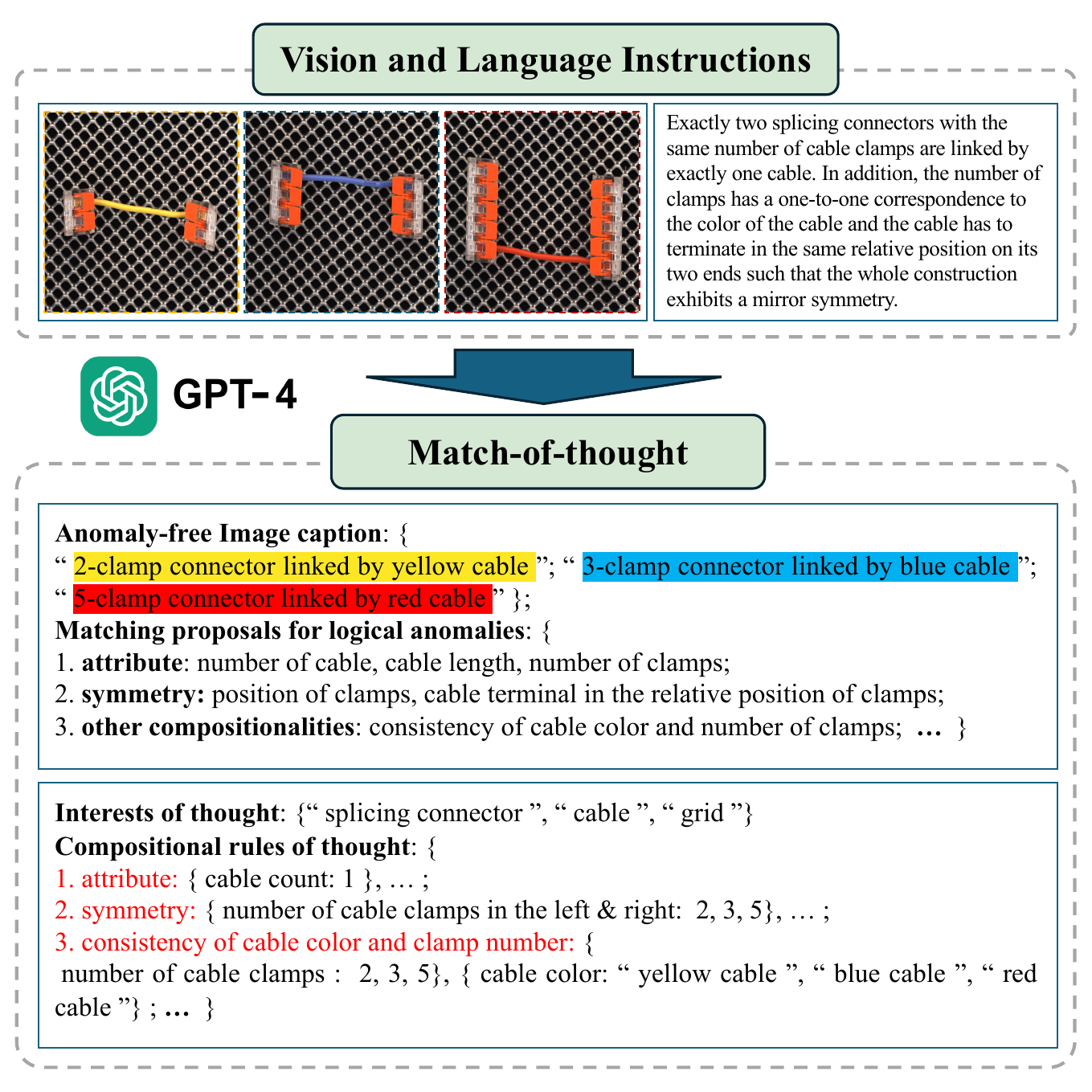} 
\caption{Match-of-thought for prompt and match engineering. The vision and language instructions consist of few anomaly-free images and compositional logical constraints in MVTec LOCO. }
\label{fig:match-of-thought}
\end{figure}

In the MVTec LOCO dataset, each category incorporates specific logical constraints.
As depicted in \cref{fig:match-of-thought}, we collect a few canonical normal images from the training set, along with their corresponding logical constraint descriptions in the original paper~\cite{bergmann2022beyond}, formulating the vision and language instructions for the MoT.
We employ the advanced GPT-4V to generate precise anomaly-free image captions and matching proposals based on the multi-modal instructions.
Subsequently, we summarize the text prompts of interests and formulate compositional matching rules.
Specifically, the MoT alleviates the issues of factuality and hallucination often encountered with LMMs in detecting logical anomalies, thereby enhancing interpretability through the effective matching of interests and compositions.
Moreover, the MoT procedure is computationally efficient as a preliminary step, making it highly adaptable to various application scenarios.

\subsection{Multi-granularity Anomaly Detectors}
Structural anomalies typically manifest as scratches or dents in localized areas, while logical anomalies involve violations of compositional constraints. 
Addressing these anomalies with a single anomaly detector is challenging.
To tackle this challenge, we propose multi-granularity anomaly detectors that aim to simultaneously detect both structural and logical anomalies within our framework.
These detectors operate at different granularities, focusing on patch, interests, and compositional matching, respectively.

\noindent\textbf{Patch Matching.} 
We systematically extract hierarchical patch features from various stages of pretrained vision backbones, including CLIP~\cite{radford2021learning} and DINOv2~\cite{oquab2023dinov2}.
Additionally, we maintain a canonical memory bank of anomaly-free image features for nearest neighbor search in patch-level anomaly detection~\cite{roth2022towards}. 
Assuming that $\bm{m} \in \mathbb{R}^{m \times d}$ and $\bm{n} \in \mathbb{R}^{n \times d}$ are $d$-dimensional patch features from the query image and memory bank respectively, the anomaly score for patch-granularity detector is defined as follows:
\begin{equation}
    s_{p} = 
    \max  (\mathbf{1}_{m \times n} - \frac{\bm{m} \cdot \bm{n}^\top}{\left\Vert \bm{m} \right\Vert  \left\Vert \bm{n} \right\Vert}),
    \label{eq:patch_score}
\end{equation}
where $\mathbf{1}_{m \times n}$ is a matrix in which all the elements are 1.

\noindent\textbf{Interest Matching.} 
Utilizing the interests of thought generated by the MoT, we combine CLIP and SAM for open-vocabulary segmentation~\cite{hajimiri2024pay, wang2024sam}, obtaining the segmentation masks of the interests. 
Subsequently, we extract sets of hierarchical interest-wise features by aggregating masked patches through average pooling.
Based on the derived feature sets of interests, we develop the interest-granularity detector, which reinterprets anomaly detection as minimum weight matching in bipartite graphs.

Let $\mathcal{P} = \{ \bm{p}_1, \bm{p}_2, \cdots, \bm{p}_i \}$ denote the feature set of $i$ interests from query image, and $\mathcal{Q} = \{ \bm{q}_1, \bm{q}_2, \cdots, \bm{q}_j \}$ denote the feature set of $j$ interests from referring anomaly-free memory bank, where $\bm{p} \in \mathbb{R}^{d}$ and $\bm{q} \in \mathbb{R}^{d}$. 
To find a bipartite matching between these two sets we search for a permutation of $N= \min(i, j)$ elements $\pi \in \mathfrak{S}_N$ with the minimal cost:
\begin{equation}
    s_{in} = \begin{cases}
    \mathop{\arg \min}\limits_{\pi \in \mathfrak{S}_N} \frac{1}{i} \sum\limits_{k=1}^{i} \mathcal{L}_{match} (\bm{p}_k, \bm{q}_{\pi(k)}), & \text{if $i \le j$,} \\
    \mathop{\arg \min}\limits_{\pi \in \mathfrak{S}_N} \frac{1}{j} \sum\limits_{k=1}^{j} \mathcal{L}_{match} (\bm{p}_{\pi(k)}, \bm{q}_k), & \text{otherwise,}
    \end{cases}
    \label{eq:interest_score}
\end{equation}
where $\mathcal{L}_{match}(\bm{p}, \bm{q})$ is a pair-wise matching cost defined as follows:
\begin{equation}
    \mathcal{L}_{match}(\bm{p}, \bm{q}) = 
    1 - \frac{\bm{p} \cdot \bm{q}^\top}{\left\Vert \bm{p} \right\Vert  \left\Vert \bm{q} \right\Vert}.
    \label{eq:cost}
\end{equation}
The optimal assignment $\pi$ can be computed efficiently using the Hungarian algorithm~\cite{kuhn1955hungarian}. 
Thus, the anomaly score for interest-granularity detector is derived from \cref{eq:interest_score}, specifically the minimal cost in bipartite matching.

\noindent\textbf{Composition Matching.} 
In the MoT, underlying logical constraints are reformulated as compositional matching rules involving visual and textural atoms.
Specifically, given a matching rule $\mathcal{R}$ that depicts the anomaly-free scenario, such as attributes or entity-relationships, relevant features of visual objects $\mathcal{V}$ and textural embeddings $\mathcal{T}$ are extracted using CLIP~\cite{radford2021learning}. These features form the essential elements to determine whether the compositional matching rule is violated. For instance, as shown in \cref{fig:framework}, for the specific matching rule $\hat{\mathcal{R}} = \{ $ ``consistency of liquid color and tag'' $\}$, we extract the relevant visual features of interests $\hat{\mathcal{V}} = \{ $``liquid in the bottle'', ``fruit''$\}$ and text embeddings $\hat{\mathcal{T}} = \{ \{ $``red liquid'', ``yellow liquid'', ``milky liquid''$\}, \{$ ``cherry'', ``orange'', ``banana''$\}\}$.
Leveraging the alignment between visual and textual features in CLIP, we can attribute the visual elements with zero-shot classification capabilities. This enables us to determine whether the visual elements violate the specified matching rule $\hat{\mathcal{R}}$.
As composition matching focuses on detecting violations of logical constraints, the anomaly score for the composition-granularity detector is defined as follows:

\begin{equation}
    s_{c} = \begin{cases}
    0, & \text{if $f(\mathcal{V}, \mathcal{T}, \mathcal{R})$ is True,} \\
    1, & \text{otherwise.}
    \end{cases}
    \label{eq:composition_score}
\end{equation}
Here, the criterion $f(\cdot)$ assesses whether interests $\langle \mathcal{V}, \mathcal{T} \rangle$ of query image conforms to the compositional rule $\mathcal{R}$.

\subsection{Calibration and Fusion}

On account that multi-granularity anomaly detectors may produce anomaly scores at different scales in measurement, it is imperative to involve score calibration and fusion modules in our framework.
With the exception of the composition-granularity detector, we calculate anomaly scores for both patch-granularity and interest-granularity detectors using statistics from anomaly-free images in the validation set.
Subsequently, the calibrated scores are standardized and passed through a sigmoid function.
The final result is determined as the maximum of the anomaly scores across the multi-granularity detectors:
\begin{equation}
    s = \max \{g(\frac{s_{p} - \mu_{p}}{\sigma_{p}}), g(\frac{s_{in} - \mu_{in}}{\sigma_{in}}), s_{c}\},
    \label{eq:final_score}
\end{equation}
where $g(\cdot)$ denotes the sigmoid function, $\mu_{p}$ and $\sigma_{p}$ represent the unbiased mean and standard deviation of patch-granularity statistics respectively, and $\mu_{in}$ and $\sigma_{in}$ denote those of interest-granularity statistics.

The overall procedure for the proposed LogSAD is summarized in \cref{alg:algorithm}.

\begin{algorithm}[tb]
\caption{LogSAD}
\label{alg:algorithm}
\textbf{Input}: Canonical normal images $\mathcal{I}_{v}$; text prompts $\mathcal{I}_{t}$; test images $\mathcal{X}$. \\
\textbf{Output}: Anomaly scores $\mathcal{S}$. 

\begin{algorithmic}[1] 
\STATE
Extract interests $\mathfrak{I}$ and compositional rules $\mathcal{R}$ via the MoT with vision and language instructions $\mathcal{I}_{v}$ and $\mathcal{I}_{t}$.

\STATE
Initialize foundation models $\mathcal{M}$. 

\STATE
Collect $\bm{n}$ and $\mathcal{Q}$ from $\mathcal{I}_{v}$ as an anomaly-free memory bank for patch and interest matching.

\FOR{$\bm{x} \in \mathcal{X}$} 
    \STATE Extract patch features $\bm{m}$ of query image $\bm{x}$, and compute patch-granularity anomaly score $s_{p}$ in \cref{eq:patch_score}.
    
    \STATE Collect feature set of interests $\mathcal{P}$ from $\bm{x}$, and compute interest-granularity anomaly score $s_{in}$ in \cref{eq:interest_score}.
    
    \STATE Extract visual and textural features $\langle \mathcal{V}, \mathcal{T} \rangle$ of $\bm{x}$, and compute compositional anomaly score $s_{c}$ in \cref{eq:composition_score}.
    
    \STATE Compute final anomaly score $s$ of $\bm{x}$ in \cref{eq:final_score}.
    
    \STATE $\mathcal{S} \gets \mathbf{UPDATE} (s) $.
    
\ENDFOR
\STATE \textbf{return} $\mathcal{S}$
\end{algorithmic}
\end{algorithm}

\section{Experiment}

\begin{table*}[ht]
\centering
\resizebox{.95\textwidth}{!}{
\begin{tabular}{cccccccc}
    \toprule
    Method & Protocol & Breakfast Box & Juice Bottle & Pushpins & Screw Bag & Splicing Connectors & Average \\
    \midrule
    SPADE~\cite{cohen2020sub}\textsuperscript{\textdagger} & \multirow{8}{*}{full-data} & 78.2 & 88.3 & 59.3 & 53.2 & 65.4 & 68.8 \\ 
    PaDim~\cite{defard2021padim} &  & 65.7 & 88.9 & 61.2 & 60.9 & 67.8 & 68.9 \\
    GCAD~\cite{bergmann2022beyond} &  & 83.9 & \textbf{99.4} & 86.2 & 63.2 & 83.9 & 83.3 \\
    PatchCore~\cite{roth2022towards}\textsuperscript{\textdagger} &  & 77.1 & 94.6 & 74.1 & 73.3 & 86.0 & 81.0 \\
    THFR~\cite{guo2023template} &  & 78.0 & 97.1 & 88.3 & 73.7 & 92.7 & 86.0  \\ 
    SINBAD~\cite{cohen2023set} &  & 92.0 & 94.9 & 78.8 & \textbf{85.4} & 92.0 & 86.8 \\
    GRAD~\cite{dai2024generating} &  & 81.2 & 97.6 & \textbf{99.7} & 76.6 & 85.4 & 87.5 \\
    GeneralAD~\cite{strater2024generalad} & & - & - & - & - & - & 84.9 \\
    \midrule
    \textbf{LogSAD (ours)}\textsuperscript{\textdagger}  & 4-shot & 94.4 & 84.3 & 82.5 & 81.5 & 88.6 & 86.3 \\
    \textbf{LogSAD (ours)}\textsuperscript{\textdagger}  & full-data & \textbf{95.7} & 95.2 & 83.6 & 83.2 & \textbf{93.5} & \textbf{90.2}  \\
    \bottomrule
\end{tabular}
}
\caption{Image-level AUROC results of unified anomaly detection on MVTec LOCO. \textsuperscript{\textdagger} indicates training-free approaches.}
\label{tab:image_auroc_on_mvtec_loco}
\end{table*}

We conduct comprehensive experiments to assess the performance of LogSAD under few-shot and full-data regimes, covering recent challenging benchmarks on industrial anomaly detection, including MVTec LOCO~\cite{bergmann2022beyond}, MVTec AD~\cite{bergmann2019mvtec} and VisA~\cite{zou2022spot} datasets.
Additionally, extensive ablation studies are conducted to validate the individual effectiveness of each proposed component.

\begin{table}[t]
\centering
\resizebox{0.95\columnwidth}{!}{
\begin{tabular}{c|c|c|c|c}
\toprule
    Method & 1-shot & 2-shot & 4-shot & Average \\
    \midrule
    PatchCore~\cite{roth2022towards}\textsuperscript{\textdagger} & 64.9 & 65.4 & 68.7 & 66.3 \\
    WinCLIP+~\cite{jeong2023winclip}\textsuperscript{\textdagger} & 68.0 & 69.7 & 71.3 & 69.7 \\ 
    PromptAD~\cite{li2024promptad} & 71.2 & 72.6 & 73.5 & 72.4 \\
    \midrule
    \textbf{LogSAD (ours)}\textsuperscript{\textdagger}  & \textbf{78.5} & \textbf{82.1} & \textbf{86.3} & \textbf{82.3} \\
    \bottomrule
\end{tabular}
}
\caption{Few shot image-level AUROC percentages of unified anomaly detection on MVTec LOCO.}
\label{tab:few_shot_results_on_mvtec_loco}
\end{table}

\begin{table}[t]
\centering
\resizebox{0.95\columnwidth}{!}{
\begin{tabular}{c|c|c|c}
\toprule
    Method & Logical & Structural & Average \\
    \midrule
    DSR~\cite{zavrtanik2022dsr} & 75.0 & 90.2 & 82.6 \\ 
    SimpleNet~\cite{liu2023simplenet} & 71.5 & 83.7 & 77.6 \\ 
    EfficientAD~\cite{batzner2024efficientad} & 86.8 & \textbf{94.7} & 90.7 \\
    \textcolor{gray}{PSAD~\cite{kim2024few}\textsuperscript{\textdaggerdbl}} & \textcolor{gray}{98.1} & \textcolor{gray}{91.6} & \textcolor{gray}{94.9} \\
    \midrule
    \textbf{LogSAD (ours)}\textsuperscript{\textdagger}  & \textbf{89.3} & 93.1 & \textbf{91.2}  \\
    \bottomrule
\end{tabular}
}
\caption{Mean anomaly detection AUROC results of detecting logical and structural anomalies respectively on MVTec LOCO. 
Note that PSAD\textsuperscript{\textdaggerdbl}, which utilizes additional segment annotations for training in anomaly detection, is not included in the comparison.
}
\label{tab:results_on_mvtec_loco_seperate}
\end{table}

\begin{table*}[t]
\centering
\resizebox{.95\textwidth}{!}{
\begin{tabular}{ccccccccccccc}
\toprule
    \multirow{4}{*}{Method} & \multicolumn{6}{c}{MVTec AD} & \multicolumn{6}{c}{VisA} \\
    \cmidrule(lr){2-7} \cmidrule(lr){8-13}
    & \multicolumn{3}{c}{image} & \multicolumn{3}{c}{pixel} & \multicolumn{3}{c}{image} & \multicolumn{3}{c}{pixel}  \\
    \cmidrule(lr){2-4} \cmidrule(lr){5-7} \cmidrule(lr){8-10} \cmidrule(lr){11-13}
    & 1-shot & 2-shot & 4-shot & 1-shot & 2-shot & 4-shot & 1-shot & 2-shot & 4-shot & 1-shot & 2-shot & 4-shot \\
    \midrule
    SPADE~\cite{cohen2020sub}\textsuperscript{\textdagger} & 81.0 & 82.9 & 84.8 & 91.2 & 92.0 & 92.7 & 79.5 & 80.7 & 81.7 & 95.6 & 96.2 & 96.6 \\ 
    PaDiM~\cite{defard2021padim} & 76.6 & 78.9 & 80.4 & 89.3 & 91.3 & 92.6 & 62.8 & 67.4 & 72.8 & 89.9 & 92.0 & 93.2 \\
    PatchCore~\cite{roth2022towards}\textsuperscript{\textdagger} & 83.4 & 86.3 & 88.8 & 92.0 & 93.3 & 94.3 & 79.9 & 81.6 & 85.3 & 95.4 & 96.1 & 96.8 \\
    WinCLIP+~\cite{jeong2023winclip}\textsuperscript{\textdagger} & 93.1 & 94.4 & 95.2 & 95.2 & 96.0 & 96.2 & 83.8 & 84.6 & 87.3 & 96.4 & 96.8 & 97.2 \\ 
    AnomalyGPT~\cite{achiam2023gpt} & 94.1 & 95.5 & 96.3 & 95.3 & 95.6 & 96.2 & 87.4 & 88.6 & 90.6 & 96.2 & 96.4 & 96.7 \\
    PromptAD~\cite{li2024promptad} & 94.6 & 95.7 & 96.6 & 95.9 & 96.2 & 96.5 & 86.9 & 88.3 & 89.1 & 96.7 & 97.1 & 97.4 \\
    \midrule
    \textbf{LogSAD (ours)}\textsuperscript{\textdagger}  & \textbf{96.1} & \textbf{96.5} & \textbf{97.0} & \textbf{97.0} & \textbf{97.3} & \textbf{97.6} & \textbf{88.2} & \textbf{90.0} & \textbf{93.0} & \textbf{97.6} & \textbf{97.8} & \textbf{98.1} \\
    \bottomrule
\end{tabular}
}
\caption{Quantitative results on MVTec AD and VisA benchmarks. Image-level and pixel-level AUROC are reported across various few-shot scenarios.}
\label{tab:quantitative_results_on_mvtec_ad_and_visa}
\end{table*}

\noindent\textbf{Datasets.} 
MVTec LOCO serves as a comprehensive benchmark designed to detect both logical and structural anomalies. 
The dataset comprises approximately 3,644 images, distributed as 1,772 images for training, 304 for validation and 1,568 for testing. 
It consists of 5 categories, \emph{i.e.} breakfast box, juice bottle, pushpins, screw bag and splicing connectors. 
The MVTec AD dataset contains 3,629 training images and 1,725 test images but pays more attention on structural anomalies than MVTec LOCO. 
It consists of 15 real-world sub-datasets, with 5 categories of textures and 10 categories of objects.
The VisA dataset consists of 9,621 normal and 1,200 anomalous color images encompassing 12 objects across 3 domains, including complex structure, single instance and multiple instances.
The anomalous images exhibit various flaws, including surface defects such as scratches, dents, color spots or cracks, and logical defects like misplacement or missing parts.

\noindent\textbf{Main Results.} 
We mainly compare our algorithm with several state-of-the-art methods on MVTec LOCO, including training-free method SPADE~\cite{cohen2020sub} and PatchCore~\cite{roth2022towards}, as well as training-based approaches such as PaDim~\cite{defard2021padim}, THFR~\cite{guo2023template}, SINBAD~\cite{cohen2023set}, GRAD~\cite{dai2024generating} and GeneralAD~\cite{strater2024generalad}, addressing both logical and structural anomalies. 
As depicted in \cref{tab:image_auroc_on_mvtec_loco}, our method achieves significant performance improvements over previous state-of-the-art methods,  particularly in terms of image-level AUROC.
As a result, our approach achieves superior results in full-data protocol without requiring training efforts, and dramatically exhibits competitive performance even in the 4-shot setting, where anomalies are detected with only 4 normal images.
Extensive experimental results demonstrate the effectiveness of our framework in simultaneously detecting both structural and logical anomalies.

In addition, we conduct experiments across extreme few-shot protocols to demonstrate the robustness of LogSAD, and provide the comparisons with advanced few-shot anomaly detection approaches, including WinCLIP~\cite{jeong2023winclip} and PromptAD~\cite{li2024promptad}. 
As shown in \cref{tab:few_shot_results_on_mvtec_loco}, our method substantially outperforms previous few-shot approaches.
Specifically, our algorithm achieves improvements of $10.3\%$, $13.1\%$ and $17.4\%$ over PromptAD under 1-shot, 2-shot and 4-shot protocols respectively, distinctly showcasing the superiority and robustness. 
Meanwhile, mean results for logical and structural anomaly detection are reported separately in \cref{tab:results_on_mvtec_loco_seperate}, along with comparisons with DSR~\cite{zavrtanik2022dsr}, SimpleNet~\cite{liu2023simplenet} and EfficientAD~\cite{batzner2024efficientad}. 
Notably, PSAD~\cite{kim2024few} is a training-based approach that utilizes additional segment annotations, making direct comparisons with other methods unfair. 
Therefore, we have excluded PSAD from our comparisons and marked it in gray in the table for clarity.

Meanwhile, quantitative results for MVTec AD and VisA benchmarks in few-shot scenarios are presented in \cref{tab:quantitative_results_on_mvtec_ad_and_visa}. 
Despite structural anomalies dominating both MVTec AD and VisA datasets, our method consistently achieves state-of-the-art performance in anomaly detection compared to previous training-free and training-based approaches, such as SPADE~\cite{cohen2020sub}, PaDim~\cite{defard2021padim}, PatchCore~\cite{roth2022towards}, WinCLIP~\cite{jeong2023winclip}, AnomalyGPT~\cite{achiam2023gpt} and PromptAD~\cite{li2024promptad}. 
In summary, extensive experimental results across various anomaly detection benchmarks demonstrate the robustness and generalization of our algorithm.
Please refer to the supplementary materials for more details.

\begin{figure}[ht]
\centering
\includegraphics[width=0.95\linewidth]{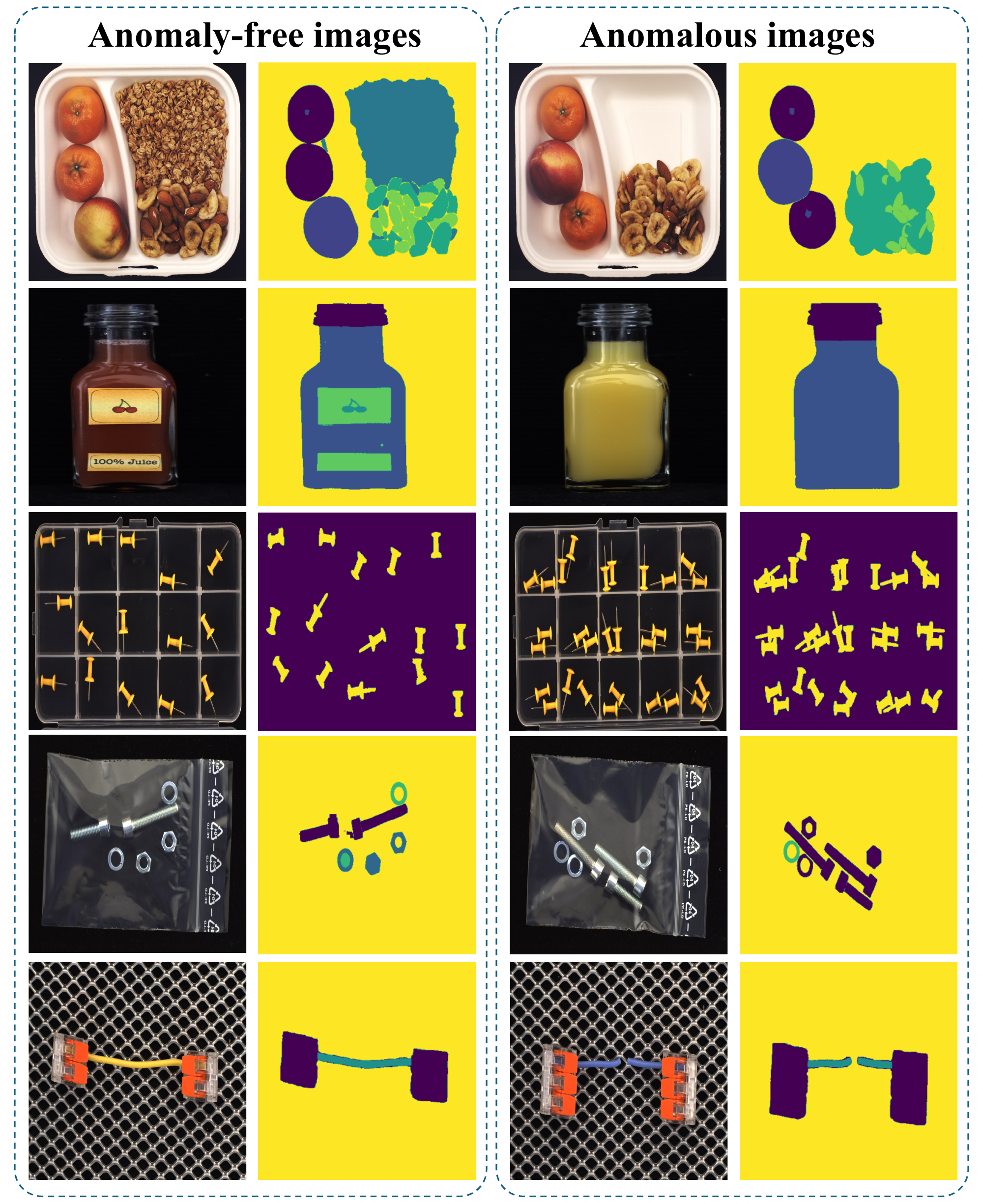} 
\caption{Qualitative visualization results of open-vocabulary semantic segmentation on MVTec LOCO. }
\label{fig:ovseg_loco}
\end{figure}

\noindent\textbf{Implementation Details.} 
Specifically, we utilize the OpenCLIP\footnote{\url{https://github.com/mlfoundations/open_clip}} implementation of CLIP and its pretrained models in our experiment. 
We employ the CLIP with ViT-L/14~\cite{dosovitskiy2020vit} pretrained on DataComp-1B~\cite{gadre2024datacomp}, DINOv2~\cite{oquab2023dinov2} with ViT-L/14, and SAM~\cite{kirillov2023segment} with ViT-H/16 as our vision and language foundation models. 
The image resolution is $448 \times 448$, and the visual feature maps extracted from ViT-L/14 across 4 hierarchical stages are upsampled to $64 \times 64$.  
Regarding pixel-level evaluation, the $64 \times 64$ anomaly maps are resized to $256 \times 256$ to ensure fair comparison with previous methodologies.

\begin{table*}[t]
\centering
\resizebox{.95\textwidth}{!}{
\begin{tabular}{c|c|c}
\toprule
    Category & Interests of Thought  & Compositional Rules of Thought  \\
    \midrule
    Breakfast Box & \makecell[c]{\{ \textbf{``orange'', ``nectarine'', ``cereals'',  ``banana chips''},  \\ \textbf{``almonds'', ``white box'',} ``black background''\} } & \makecell[c]{ \{``the ratio and relative position of the cereals, \\ banana chips and almonds should be fixed'' \} } \\ 
    \midrule
    Juice Bottle & \makecell[c]{\{ \textbf{``glass'', ``liquid in the bottle'', ``fruit''}, \\ \textbf{``label'',} ``black background''\}  }  & \{\textbf{``consistency of fruit tag and liquid color''} \}  \\ 
    \midrule
    Pushpins & \{ \textbf{``pushpin'',} ``plastic box'', ``black background''\} & \{\textbf{``number of pushpins is 15''} \} \\
    \midrule
    Screw Bag & \makecell[c]{\{ \textbf{``screw'', ``hex nut'', ``ring washer'',} \\ ``plastic bag'', ``background''\}  } & \makecell[c]{\{\textbf{``histogram of screws''}, \\ ``1 long screw, 1 short screw, 2 nuts and 2 washers'' \}} \\
    \midrule
    Splicing Connectors & \{ \textbf{``splicing connector'', ``cable'',} ``grid''\} & \makecell[c]{\{\textbf{``consistency of cable color and number of clamps'', ``number of cable is 1'',} \\ \textbf{``splicing connectors consist of left and right parts and keep symmetry''},  \\ ``cable terminates in the same relative position''\}} \\
    \bottomrule
\end{tabular}
}
\caption{Interests and compositional rules of thought on MVTec LOCO. 
The final adopted interests and compositional rules for matching are marked in \textbf{bold}.
}
\label{tab:interests_of_thought}
\end{table*}

\noindent\textbf{Matching Details.} 
We explicitly extract normalized hierarchical features from the 4 stages of ViT-L/14.
The ensemble score maps across these stages are then averaged for patch matching.
In the full-data setting, the memory bank of normal images is downsampled via greedy coreset subsampling to reduce redundancy~\cite{roth2022towards}.
In addition, detailed interests and compositional rules for each category on MVTec LOCO are presented in \cref{tab:interests_of_thought}.
By default, we use the text prompt templates designed for ImageNet in OpenAI's CLIP~\cite{radford2021learning} and average the predictions across multiple text prompts with these templates.
Note that the background classes of interests (\emph{e.g.} ``black background'', ``plastic bag'' and ``grid'') are introduced to improve accurate open-vocabulary semantic segmentation.
In practice, we select the foreground classes of interests and specific compositional rules for matching, excluding prompts that are ambiguous or have limited performance in open-vocabulary semantic segmentation.
Subsequently, masked visual tokens are aggregated using average pooling for zero-shot classification, leveraging aligned VLMs such as CLIP.
This method aids in distinguishing the fine-grained categories or attributes in compositional rules with masked features, \emph{e.g.} ``red cable, blue cable or yellow cable'', ``cherry, banana or orange tag''.

\noindent\textbf{Ablation Studies.} 
Given the pivotal role of defining interests in our framework, we present the intermediate segmentation results of interests necessary for collecting interest sets used in bipartite graph matching.
Moreover, precise open-vocabulary segmentation results facilitate the implementation of compositional rules involving visual and textural elements.
As shown in \cref{fig:ovseg_loco}, the impressive visualizations highlight the capability to achieve high-quality segmentation results using the MoT and VLMs, thereby eliminating the necessity for training efforts.

\begin{table}[t]
\centering

\resizebox{0.95\columnwidth}{!}{
\begin{tabular}{c|c|c|c|c}
\toprule
\multirow{2}{*}{Detectors} & \multicolumn{2}{c}{full-data} & \multicolumn{2}{c}{4-shot} \\
\cmidrule(lr){2-3} \cmidrule(lr){4-5}
& Structural & Logical & Structural & Logical \\
\midrule
Patch  & \textbf{93.1} & 71.8 & \textbf{87.3} & 67.6 \\
Insterests  & 81.7 & \textbf{80.6} & 72.1 & 74.3 \\
Composition  & 58.0 & 78.2 & 58.0 & \textbf{78.2} \\
\bottomrule
\end{tabular}
}
\caption{Ablation studies on multi-granularity detectors. Image-level AUROC results of respective detectors in detecting structural and logical anomalies under full-data and 4-shot protocols are reported on MVTec LOCO.}
\label{tab:ablation_on_detectors}
\end{table}

\begin{table}[t]
\centering

\resizebox{.9\columnwidth}{!}{
\begin{tabular}{ccccc}
\toprule
Patch & Interests & Composition & $F_1$-max & AUROC \\
\midrule
\checkmark & &  & 81.3 & 81.0  \\
& \checkmark & & 81.7 & 80.7 \\
& \checkmark & \checkmark & 83.8 & 85.1 \\
\checkmark & \checkmark &  & 85.5 & 86.4 \\
\checkmark & \checkmark & \checkmark & \textbf{88.8} & \textbf{90.2}  \\
\bottomrule
\end{tabular}
}
\caption{Ablation studies on detectors calibration and fusion. Image-level $F_1$-max and AUROC results are presented on MVTec LOCO.}
\label{tab:ablation_on_mvtec_loco}
\end{table}

In addition, we conduct ablation studies on MVTec LOCO to highlight the importance of multi-granularity anomaly detectors. 
As demonstrated in \cref{tab:ablation_on_detectors}, multi-granularity anomaly detectors prove effective and advantageous in various respects.
For instance, the patch-granularity detector focuses on structural anomalies, the composition-granularity detector addresses logical anomalies with compositionality, and the interest-granularity detector achieves a balance between detecting both structural and logical anomalies.
Furthermore, both the patch-granularity and interest-granularity detectors show significant improvement as the number of normal images increases.
The composition-granularity detector operates independently of reference images, ensuring stability in detecting logical anomalies across both full-data and few-shot protocols.

Notably, we highlight the image-level $F_1$-max and AUROC scores of individual detectors and their fusion for unified anomaly detection in \cref{tab:ablation_on_mvtec_loco}. 
The experimental results demonstrate how different anomaly detectors complement each other in detecting both structural and logical anomalies, underscoring the importance of multi-granularity detectors.
Furthermore, the ensemble of multi-granularity detectors dramatically enhances the performance on anomaly detection, highlighting the effectiveness of our calibration and fusion strategy.

\begin{figure}[t]
\centering
\includegraphics[width=0.95\linewidth]{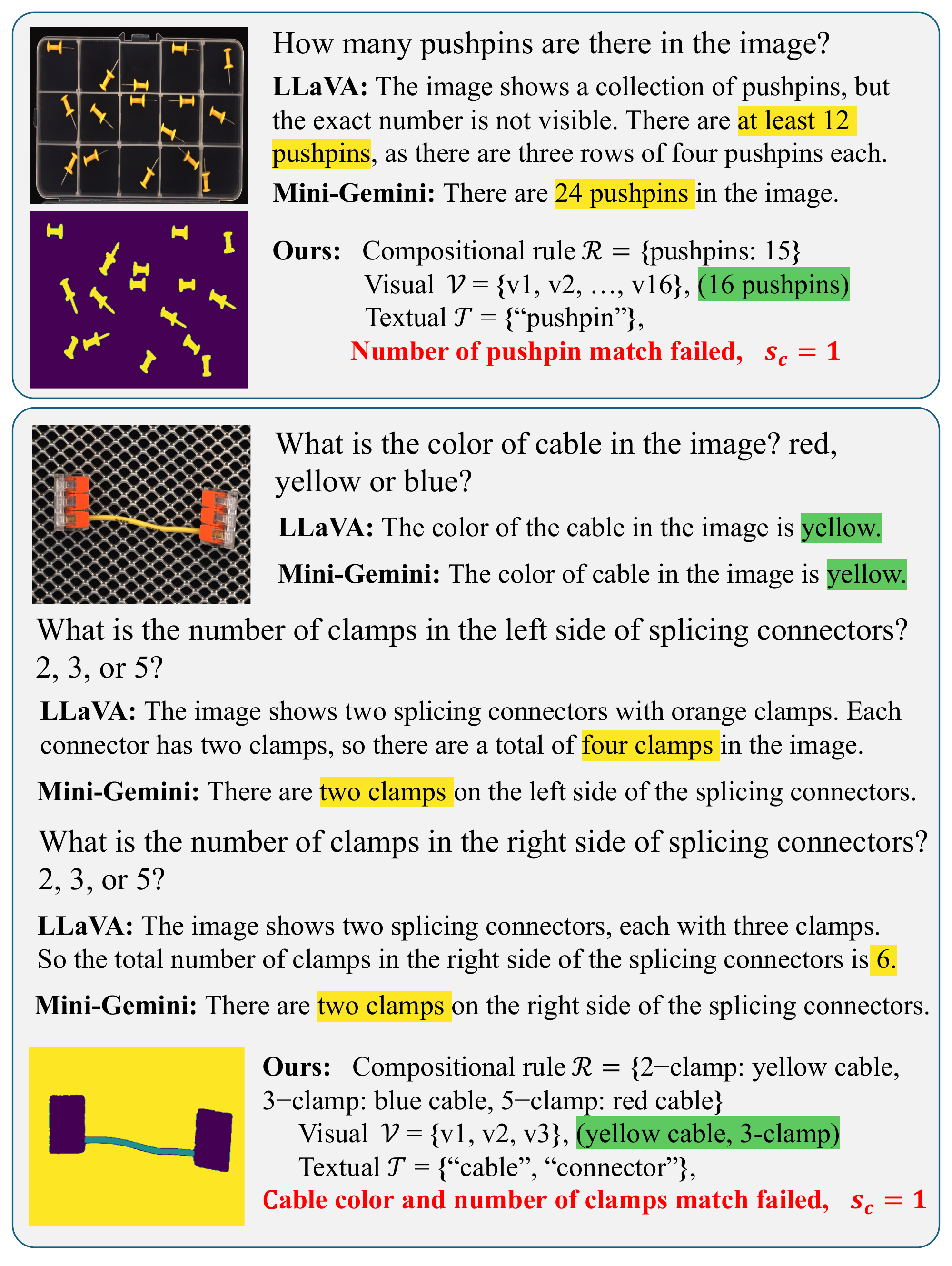} 
\caption{Comparison with LMMs in compositionality. 
Correct answers are marked in green, while incorrect ones are marked in yellow.
Our proposed composition-granularity detector performs effectively, whereas LMMs struggle with issues of factuality and hallucination in logical anomaly understanding and detection.}
\label{fig:comparison_with_lmms}
\end{figure}

\noindent\textbf{Comparison with LMMs in compositional anomalies.} 
Recent empirical studies~\cite{thrush2022winoground, yuksekgonul2023and, mitra2024compositional, zeng2024investigating} have shown that LMMs struggle to perform compositional visual understanding, particularly in tasks involving identifying object attributes and inter-object relations.
Specifically, studies indicate that VLMs~\cite{ma2023crepe} often learn a ``bag of objects" representation, which limits their compositional capabilities.
Here, we assess the compositionality performance of advanced LMMs, such as LLaVA~\cite{liu2024visual, liu2024improved} and Mini-Gemini~\cite{li2024mini}. 
As illustrated in \cref{fig:comparison_with_lmms}, LMMs provide accurate responses in specific tasks, such as identifying the color of cables in splicing connectors.
However, LMMs face challenges in tasks requiring complex scene understanding, including object counting and spatial perception, \emph{e.g.} counting the number of pushpins or distinguishing the left and right sides of clamps.
Consequently, LMMs continue to exhibit issues such as hallucinations, bias, and factual inaccuracies, making them impractical for addressing most logical anomalies involving compositionality.

In contrast, the proposed composition-granularity detector effectively handles the compositionality using visual and textural atoms, leveraging open-vocabulary segmentation masks.
Moreover, it is worth noting that the composition-granularity detector effectively distinguishes various fine-grained logical anomalies.
With aligned visual and textual embeddings from CLIP, it is available to extract relevant visual objects or text descriptions for attributes in compositional matching, \emph{e.g.}, the number of pushpins, cable color, and the count of clamps in splicing connectors, and so on.
Quantitative experiment results in \cref{tab:ablation_on_detectors} and \cref{tab:ablation_on_mvtec_loco} demonstrate the effectiveness of our composition matching approach in addressing logical anomalies and its contribution to unified anomaly detection.

\section{Conclusion}

In this paper, we propose LogSAD, a unified multi-modal framework for detecting both structural and logical anomalies without training endeavors. 
We elaborate on multi-granularity anomaly detectors designed to detect anomalies across diverse granularities by leveraging vision and language foundation models. 
Subsequently, anomaly scores from different anomaly detectors are calibrated and fused for the final decision procedure.
Extensive experiments are conducted across various anomaly detection datasets to demonstrate the effectiveness of our method.

However, our approach is not without limitations. 
Future work will focus on enhancing the performance of open-vocabulary semantic segmentation and exploring advanced LMMs for complicated scenarios.

\section{Acknowledgement}

This work is partly supported by the National Key Research and Development Plan (2024YFB3309302).

{
    \small
    \bibliographystyle{ieeenat_fullname}
    \bibliography{main}
}


\clearpage
\setcounter{page}{1}
\maketitlesupplementary

In this supplementary material, we provide details of  experimental settings including data preprocessing and evaluation metrics.
Additionally, quantitative results on MVTec LOCO~\cite{bergmann2022beyond}, MVTec AD~\cite{bergmann2019mvtec} and VisA~\cite{zou2022spot} benchmarks are presented to demonstrate the effectiveness of our algorithm.
Finally, we provide a comprehensive analysis and discussion of our framework.

\section{Experimental Details}

\noindent\textbf{Data Preprocessing.} 
Regarding vision and language foundation models including CLIP~\cite{radford2021learning}, DINOv2~\cite{oquab2023dinov2} and SAM~\cite{kirillov2023segment}, we apply the same data preprocessing pipeline across MVTec LOCO, MVTec AD and VisA datasets to mitigate potential train-test discrepancy. 
Specifically, it involves channel-wise standardization with the pre-computed mean $[0.48145466, 0.4578275, 0.40821073]$ and standard deviation $[0.26862954, 0.26130258, 0.27577711]$ after normalizing each RGB image into $[0, 1]$, followed by bicubic interpolation based on the Pillow implementation.

\noindent\textbf{Evaluation Metrics.} 
Consistent with existing methods~\cite{bergmann2019mvtec, bergmann2022beyond}, we report the results of the Area Under the Receiver Operator Curve (AUROC) documented in the body of the paper for the evaluation of image-level anomaly detection and pixel-level anomaly localization.
Additionally, we supplement the $F_1$-max results in anomaly detection.
The $F_1$-max score is computed from the precision and recall for the anomalous samples at the optimal threshold, which is a more straightforward metric to measure the upper bound of anomaly prediction performance across thresholds.

\section{Quantitative Results}

To elucidate the interaction between patch matching and composition matching in detecting logical and structural anomalies, we present the experimental results in \cref{tab:r2}, demonstrating that incorporating composition matching with patch matching improves results for logical AD and achieves comparable results for structural AD. 
Quantitative results indicate that the inclusion of composition matching significantly enhances detection performance for logical anomalies while maintaining comparable performance on structural anomalies.
Additionally, we report the detailed subset-level results of LogSAD. 
Specifically, the results on MVTec LOCO~\cite{bergmann2022beyond} are presented in \cref{tab:quantitative_results_on_mvtec_loco}, and the results on MVTec AD~\cite{bergmann2019mvtec} and VisA~\cite{zou2022spot} benchmarks are depicted in \cref{tab:quantitative_results_on_mvtec_ad} and \cref{tab:quantitative_results_on_visa}, respectively.

\begin{table}[ht]
\centering
\resizebox{0.9\columnwidth}{!}{
\begin{tabular}{c|c|c|c}
\toprule
Detectors & Structural & Logical & Average \\
\midrule
Patch  & 87.3 & 67.6 & 77.4 \\
Composition  &  58.0 & 78.2 & 68.1 \\
Patch + Composition  & 85.8 & 82.0 & 83.9 \\
\bottomrule
\end{tabular}
}
\caption{Image-level AUROC of multi-granularity detectors under 4-shot protocol on MVTec-LOCO dataset.}
\label{tab:r2}
\end{table}

\section{Discussion}

\noindent\textbf{Canonical Normal Images in the MoT.} 
The MVTec LOCO dataset~\cite{bergmann2022beyond} contains a varying number of normal sub-classes in each category, \eg 1, 3, 1, 1, 3 corresponding to ``breakfast box", ``juice bottle", ``pushpins", ``screw bag", and ``splicing connectors". 
Specifically, the "juice bottle" category includes cherry, banana, and orange labels, while the "splicing connectors" category contains red, blue, and yellow cables.
Thus, we sample 3 canonical normal images from the normal sub-classes in each category to maximize sub-class coverage in the training set, facilitating the establishment of comprehensive matching interests and compositional rules.
We typically observe that the generated results from GPT-4V remain consistent across the provided normal images due to subtle visual variations within each sub-class.
Notably, the sampled canonical normal images and GPT-4V are exclusively used for offline proposal generation, which operates independently of the anomaly detection algorithm.

In practice, the quality of generated proposals can be assessed in various perspectives, including the qualitative results of open-vocabulary semantic segmentation in terms of interests of thought, as depicted in \cref{fig:ovseg_loco}, and quantitative results through interest matching and compositional matching, as shown in \cref{tab:ablation_on_detectors}.

\begin{figure}[t]
\centering
\includegraphics[width=0.48\textwidth]{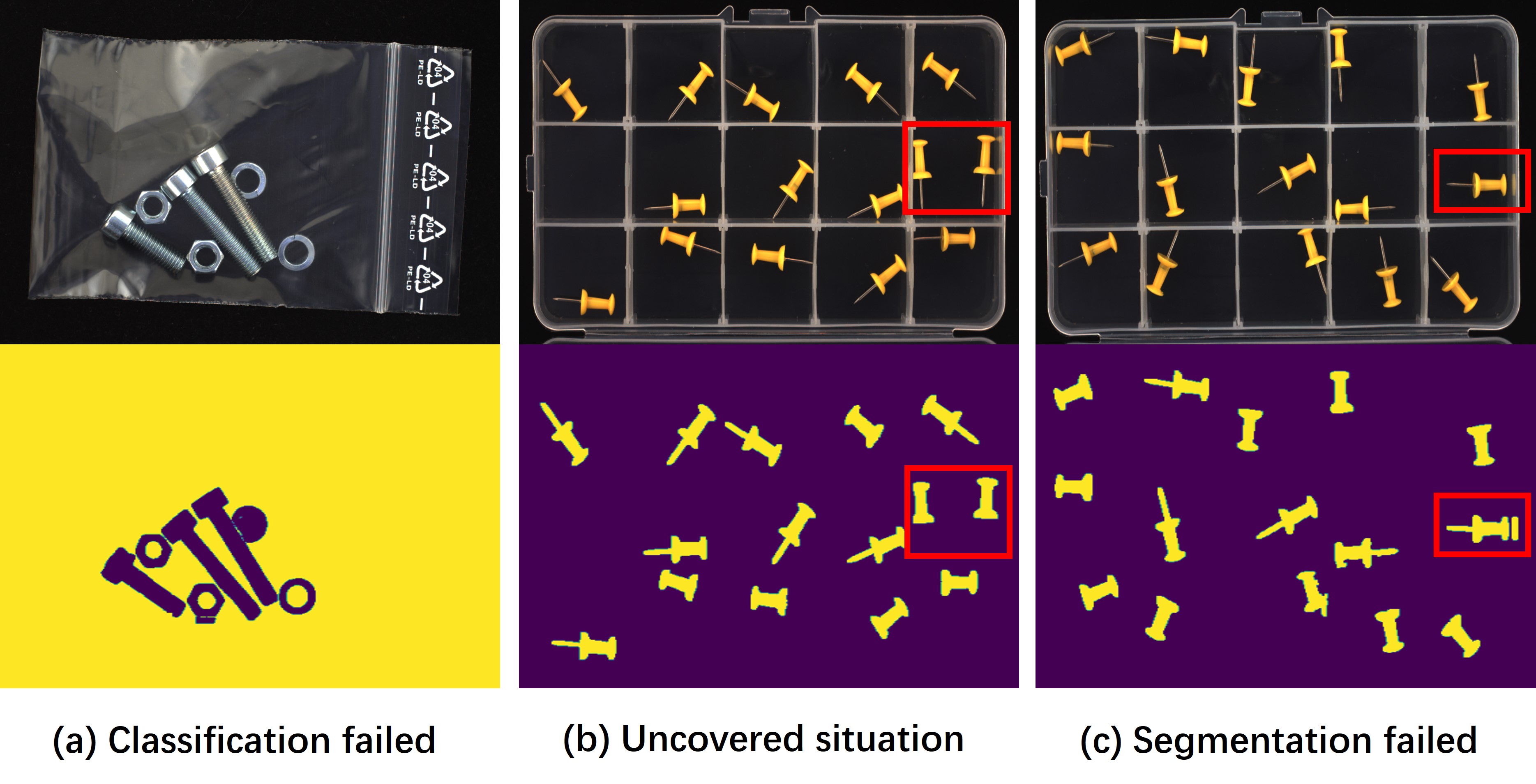} 
\caption{Failure cases of LogSAD. }
\label{fig:failure_cases}
\end{figure}

\noindent\textbf{Failure cases.} 
In addition, we present the failure cases in \cref{fig:failure_cases} to address the limitations of our framework, including failures in open-vocabulary semantic segmentation and uncovered situations in compositional matching. 
For instance, (a) fails to distinguish the ``hex nut'' and the ``ring washer''; 
(b) the number of pushpins is 15, but two pushpins appear in one division, which is not covered by matching rules;
(c) fails in open-vocabulary semantic segmentation and counting due to the reflection of pushpins.

\noindent\textbf{Computation Analysis.} 
Previous methods, such as WinCLIP~\cite{jeong2023winclip}, PromptAD~\cite{li2024promptad}, AnomalyCLIP~\cite{zhou2024anomalyclip} fine-tuning with CLIP, and AnomalyGPT~\cite{gu2024anomalygpt} fine-tuning with larger models (\eg Vicuna-7B and Vicuna-13B), focus solely on structural AD but continue to struggle with logical AD.
Our focus is on the training-free application of off-the-shelf foundation models for both logical and structural AD. 
Note that we use GPT-4V only for offline match proposal generation, and open-sourced foundation models including CLIP, DINOv2 and SAM are collaborated for anomaly detection with around 1.3B parameters.
Consequently, \cref{tab:quantitative_results_on_mvtec_ad_and_visa} shows our method achieves state-of-the-art performance on structural AD datasets, outperforming training-based methods like PromptAD and AnomalyGPT.
In addition, experimental results in \cref{tab:image_auroc_on_mvtec_loco}, \cref{tab:few_shot_results_on_mvtec_loco} and \cref{tab:results_on_mvtec_loco_seperate} demonstrate the effectiveness of our framework in both logical and structural AD.

\begin{table*}[ht]
\centering
\resizebox{1.\textwidth}{!}{
\begin{tabular}{ccccccccccccc}
    \toprule
    \multirow{2}{*}{Protocol} & \multicolumn{2}{c}{Breakfast Box} & \multicolumn{2}{c}{Juice Bottle} & \multicolumn{2}{c}{Pushpins} & \multicolumn{2}{c}{Screw Bag} & \multicolumn{2}{c}{Splicing Connectors} & \multicolumn{2}{c}{\textbf{Average}} \\
    \cmidrule(lr){2-3} \cmidrule(lr){4-5} \cmidrule(lr){6-7} \cmidrule(lr){8-9} \cmidrule(lr){10-11} \cmidrule(lr){12-13}
    & $F_1$-max & AUROC & $F_1$-max & AUROC & $F_1$-max & AUROC & $F_1$-max & AUROC & $F_1$-max & AUROC & $F_1$-max & AUROC \\
    \midrule
    1-shot & 85.0 & 88.0 & 85.6 & 78.1 & 75.7 & 78.0 & 80.7 & 70.6 & 78.2 & 77.7 & 81.0 & 78.5 \\
    2-shot & 88.1 & 91.5 & 85.7 & 77.5 & 77.8 & 81.1 & 83.0 & 80.5 & 78.2 & 79.8 & 82.6 & 82.1 \\
    4-shot & 89.9 & 94.4 & 88.2 & 84.3 & 81.4 & 82.5 & 84.1 & 81.5 & 84.7 & 88.6 & 85.7 & 86.3 \\
    full-data & 92.0 & 95.7 & 94.0 & 95.2 & 81.3 & 83.6 & 85.2 & 83.2 & 91.3 & 93.5 & 88.8 & 90.2  \\ 
    \bottomrule
\end{tabular}
}
\caption{Image-level $F_1$-max and AUROC results on MVTec LOCO in few-shot and full-data protocols.}
\label{tab:quantitative_results_on_mvtec_loco}
\end{table*}

\begin{table*}[ht]
\centering
\resizebox{1.\textwidth}{!}{
\begin{tabular}{ccccccccccccc}
    \toprule
    \multirow{3}{*}{Category} & \multicolumn{6}{c}{image} & \multicolumn{6}{c}{pixel}  \\ 
    \cmidrule(lr){2-7} \cmidrule(lr){8-13} 
    & \multicolumn{2}{c}{1-shot} & \multicolumn{2}{c}{2-shot} & \multicolumn{2}{c}{4-shot} & \multicolumn{2}{c}{1-shot} & \multicolumn{2}{c}{2-shot} & \multicolumn{2}{c}{4-shot} \\
    \cmidrule(lr){2-3} \cmidrule(lr){4-5} \cmidrule(lr){6-7} \cmidrule(lr){8-9} \cmidrule(lr){10-11} \cmidrule(lr){12-13}
    & $F_1$-max & AUROC & $F_1$-max & AUROC & $F_1$-max & AUROC & $F_1$-max & AUROC & $F_1$-max & AUROC & $F_1$-max & AUROC \\
    \midrule
    bottle & 100 & 100 & 100 & 100 & 100 & 100 & 81.5 & 99.0 & 82.1 & 99.1 & 82.8 & 99.2 \\
    cable & 88.0 & 90.4 & 87.9 & 91.2 & 87.8 & 90.8 & 60.7 & 96.7 & 62.7 & 97.4 & 63.1 & 97.6 \\
    capsule & 94.0 & 92.0 & 94.9 & 92.7 & 97.3 & 94.1 & 50.3 & 98.0 & 50.5 & 98.3 & 51.7 & 98.4 \\
    carpet & 98.9 & 99.4 & 98.9 & 99.3 & 98.9 & 99.4 & 67.6 & 99.2 & 67.4 & 99.2 & 67.5 & 99.2 \\ 
    grid & 99.1 & 99.8 & 100 & 100 & 100 & 100 & 51.2 & 99.3 & 55.9 & 99.5 & 55.9 & 99.5 \\
    hazelnut & 99.3 & 99.9 & 98.6 & 99.8 & 100 & 100 & 65.6 & 98.9 & 67.7 & 99.1 & 71.5 & 99.3 \\
    leather & 99.5 & 99.9 & 99.5 & 99.9 & 100 & 100 & 49.4 & 99.3 & 48.0 & 99.3 & 48.9 & 99.4 \\
    metal\_nut & 99.5 & 99.6 & 99.5 & 99.7 & 100 & 100 & 74.7 & 96.0 & 76.7 & 96.3 & 84.6 & 97.8 \\
    pill & 96.2 & 91.1 & 96.4 & 97.2 & 96.8 & 97.9 & 66.7 & 96.8 & 67.7 & 97.0 & 68.4 & 97.1 \\
    screw & 92.2 & 92.4 & 92.2 & 92.4 & 92.2 & 92.4 & 18.7 & 95.6 & 22.3 & 96.6 & 27.5 & 97.4 \\
    tile & 98.8 & 99.9 & 100 & 100 & 100 & 100 & 71.6 & 96.3 & 72.1 & 96.5 & 72.3 & 96.6 \\
    toothbrush & 92.9 & 93.9 & 91.8 & 92.8 & 93.3 & 92.2 & 38.8 & 96.2 & 38.1 & 96.2 & 37.5 & 96.1 \\
    transistor& 79.5 & 89.5 & 78.5 & 88.5 & 78.7 & 90.9 & 48.9 & 90.4 & 50.9 & 91.7 & 52.2 & 91.9 \\
    wood & 99.2 & 99.8 & 99.2 & 99.7 & 99.2 & 99.8 & 70.2 & 97.0 & 70.2 & 97.0 & 70.2 & 97.0 \\
    zipper & 96.8 & 93.5 & 98.3 & 94.7 & 99.2 & 97.6 & 56.1 & 96.6 & 58.2 & 97.1 & 58.6 & 97.3 \\
    average & 95.6 & 96.1 & 95.7 & 96.5 & 96.2 & 97.0 & 58.1 & 97.0 & 59.4 & 97.3 & 60.8 & 97.6 \\
    \bottomrule
\end{tabular}
}
\caption{Image-level/pixel-level $F_1$-max and AUROC results on MVTec AD.}
\label{tab:quantitative_results_on_mvtec_ad}
\end{table*}

\begin{table*}[ht]
\centering
\resizebox{1.\textwidth}{!}{
\begin{tabular}{ccccccccccccc}
    \toprule
    \multirow{3}{*}{Category} & \multicolumn{6}{c}{image} & \multicolumn{6}{c}{pixel}  \\ 
    \cmidrule(lr){2-7} \cmidrule(lr){8-13} 
    & \multicolumn{2}{c}{1-shot} & \multicolumn{2}{c}{2-shot} & \multicolumn{2}{c}{4-shot} & \multicolumn{2}{c}{1-shot} & \multicolumn{2}{c}{2-shot} & \multicolumn{2}{c}{4-shot} \\
    \cmidrule(lr){2-3} \cmidrule(lr){4-5} \cmidrule(lr){6-7} \cmidrule(lr){8-9} \cmidrule(lr){10-11} \cmidrule(lr){12-13}
    & $F_1$-max & AUROC & $F_1$-max & AUROC & $F_1$-max & AUROC & $F_1$-max & AUROC & $F_1$-max & AUROC & $F_1$-max & AUROC \\
    \midrule
    candle & 88 & 92.5 & 86.7 & 92.0 & 87.4 & 92.4 & 36.2 & 98.2 & 36.5 & 98.9 & 36.2 & 98.9 \\
    capsules & 91.9 & 96.0 & 92.5 & 96.5 & 93.1 & 97.1 & 40.3 & 96.6 & 42.0 & 96.8 & 44.9 & 97.8 \\
    cashew & 81.2 & 78.2 & 84.2 & 83.5 & 91.5 & 93.7 & 62.8 & 98.5 & 63.0 & 98.6 & 62.8 & 98.5 \\
    chewinggum & 95.0 & 97.7 & 96.0 & 97.3 & 97.5 & 98.7 & 69.6 & 99.5 & 70.1 & 99.6 & 69.4 & 99.5 \\ 
    fryum & 91.5 & 93.7 & 92.8 & 96.0 & 96.5 & 98.3 & 33.7 & 93.9 & 38.7 & 95.0 & 41.8 & 95.1 \\
    macaroni1 & 84.7 & 89.6 & 85.1 & 90.9 & 90.1 & 93.7 & 27.1 & 98.6 & 29.5 & 98.9 & 29.0 & 99.1 \\
    macaroni2 & 69.2 & 68.3 & 68.7 & 68.5 & 72.7 & 75.9 & 14.4 & 97.9 & 13.2 & 98.1 & 16.9 & 98.3 \\
    pcb1 & 85.3 & 91.3 & 85.1 & 91.8 & 84.3 & 91.3 & 52.0 & 98.0 & 48.7 & 98.2 & 48.5 & 98.2 \\
    pcb2 & 79.1 & 84.6 & 80.5 & 86.5 & 80.4 & 87.3 & 36.1 & 98.0 & 36.7 & 98.0 & 37.8 & 98.2 \\
    pcb3 & 76.2 & 82.3 & 81.9 & 87.6 & 87.9 & 93.5 & 40.6 & 97.6 & 38.8 & 98.1 & 40.4 & 98.5 \\
    pcb4 & 82.6 & 84.9 & 85.5 & 89.3 & 89.9 & 94.7 & 39.1 & 94.8 & 40.8 & 94.5 & 43.6 & 96.0 \\
    pipe\_fryum & 98.0 & 99.5 & 98.0 & 99.7 & 98.0 & 99.5 & 59.2 & 99.1 & 58.2 & 99.1 & 60.3 & 99.2 \\
    average & 85.2 & 88.2 & 86.4 & 90.0 & 89.1 & 93.0 & 42.6 & 97.6 & 43.0 & 97.8 & 44.3 & 98.1 \\
    \bottomrule
\end{tabular}
}
\caption{Image-level/pixel-level $F_1$-max and AUROC results on VisA.}
\label{tab:quantitative_results_on_visa}
\end{table*}

\end{document}